\documentclass{article}

\usepackage[utf8]{inputenc} 
\usepackage[T1]{fontenc}    
\usepackage{hyperref}       
\usepackage{url}            
\usepackage{booktabs}       
\usepackage{amsfonts}       
\usepackage{microtype}      
\usepackage[]{algorithm2e}
\usepackage{amsmath}
\usepackage{graphicx}
\usepackage{caption}
\usepackage{subcaption}
\usepackage[]{natbib}

\renewcommand{\AA}{\mathbb{A}}
\renewcommand{\SS}{\mathbb{S}}

\newcommand{\PP}{\mathcal{P}}
\newcommand{\tp}[1]{\langle{#1}\rangle}

\title{Practical Issues of Action-conditioned \\ Next Image Prediction}
\author{
{\small Donglai Zhu\thanks{These two authors contributed equally.}, Hao Chen\footnotemark[1], Hengshuai Yao, Masoud Nosrati, Peyman Yadmellat, Yunfei Zhang}\\
\texttt{\footnotesize donglai.zhu,hao.chen1,hengshuai.yao,masoud.nosrati,peyman.yadmellat,yunfei.zhang1@huawei.com} \\
 Huawei Research Canada \\
}

\begin{document}

\maketitle

\begin{abstract}
The problem of action-conditioned image prediction in robotics is to predict the expected next frame given the current camera frame the robot observes and the action it selects.  We provide the first comparison of two recent popular models, Convolutional Dynamic Neural Advection (CDNA) \citep{video-predict-goodfellow-16} and a feedforward model \citep{action_video_prediction_honglak}, especially for image prediction on cars. Our major finding is that action tiling encoding is the most important factor leading to the remarkable performance of the CDNA model. We present a light-weight model by action tiling encoding which has a single-decoder feedforward architecture same as \cite{action_video_prediction_honglak}. On a real driving dataset, the CDNA model achieves ${0.3986} \times 10^{-3}$ MSE and ${0.9836}$ Structure SIMilarity (SSIM) with a network size of about { ${12.6}$ million} parameters. With a small network of fewer than { ${{1}}$ million}  parameters, our new model achieves a comparable performance to CDNA at ${0.3613} \times 10^{-3}$ MSE and ${0.9633}$ SSIM. Our model requires less memory, is more computationally efficient and more advantageous to be used inside self-driving vehicles. 

\end{abstract}



\renewcommand{\cite}{\citep}
\section{Introduction}
In autonomous driving and robotics, system engineers often choose a state representation according to the application. For different control problems, the representation is usually different. It is also common that for the same control problem, different algorithms use different state representations. In particular, we are interested in models that reflect how a vehicle's projection of the world changes according to its action. 

In the {classical approach} for autonomous driving, engineers extract perception and localization results from sensors, summarizing the geometric relationship of the vehicle with its environment. Based on the {\em geometric representation} of the world, a controller is built. This approach is so far the most popular and widely adopted in industry.  
However, it also has weakness. The driving software following this approach has a heavy pipeline stack, from perception and localization to planning modules.
The message passing between perception, localization and controller modules leads to a complicated software architecture. On the control side it is common to use heuristics and hand-tuned controllers that only know how to respond to a low-dimensional state representation.  This specific design for autonomous driving architecture results in significant software development efforts, requiring intensive testing efforts for corner cases and tuning efforts for hyper parameters. By following this classical approach, building safety has been largely reduced to software practice.   

{The imitation learning approach}, starting from multi-layer perceptions (MLPs) \citep{alvinn} to the more recent convolutional neural networks \cite{nvidia-2016}, regresses the steering angle given the camera view.
This approach leads to a simpler architecture for autonomous driving. However, comparing to the classical approach, it lacks understanding of the decision process of driving. The effectiveness of this approach is yet to be seen in the next few years. 

{The affordance approach}\cite{DeepDriving-2015} predicts relevant geometric features (called ``affordances'') from images. Based on the predicted features, a controller can be developed. This approach bears some similarity to Pavlovian control in which animals map predictions of events into their behaviors\citep{modayil_palovian}.

A key difference of the three approaches is the {state representation} on which controller is built. The classical approach represents the world as geometric features either manually extracted or learned using machine learning. The end-to-end imitation learning approach uses raw images as the world representation. The affordance approach can be viewed as an intermediate approach between the two: it uses {predicted geometric information} from images as the state representation.   

These are exciting advances on the control side of autonomous vehicles. An autonomous vehicle needs to drive safely but it also needs to be understood by human beings. It has to keep the boarding passengers well informed during the planning and decision making so that they can trust the system and alleviate unnecessary nervousness. In addition, system designers need to understand the decision making process of the software so it can be improved.  
However, by far visualization of the internal planning and decision making process of autonomous driving softwares has remained less noticed. 
Given the many problems that self-driving cars are facing and their rich choices of state representation, it is important to model the effects of car actions in a representation-agnostic way. 
Models that estimate environment state transitioning in response to a robot's actions turn out to be fundamental for visualizing the planning process of self-driving cars. 
   
Besides visualization, building action models is essential to bring reinforcement learning to autonomous driving. Reinforcement learning achieved remarkable success in Atari games \cite{mnih-dqn-2015} and Go \cite{go-nature-2016}. 
Using reinforcement learning to develop driving software has a potential of making cars safer and save human lives.  
Lacking an accurate action model is one of the major gaps from reinforcement learning to autonomous driving:
A perfect simulator or action model such as in games is not available in driving. A perfect simulator means ground truth interaction samples can be collected efficiently, cheaply and infinitely. AlphaGo achieves master play through playing millions of games against itself, state-of-art Go programs and human professionals \citep{go-nature-2016}. Recently AlphGo Zero even beat its previous version by learning through playing itself without the use of human knowledge \citep{alphaGo0-2017}. For cars to be able to gain driving skills off-line like AlphaGo, we need  models that accurately predict transitioning between states conditioned on actions. Previous works in robotics and autonomous driving have focused on modeling vehicle dynamics \citep{omar1998vehicle,yim2004modeling,ng2003autonomous,bakker1987tyre,kong2015kinematic, rajamani2011vehicle,levinson2011towards,urmson2007tartan}.  

Learning action models for driving is challenging.  For a contrast, state transition in the game of Go is deterministic and noise-free. In Go, given a board and a legal move the next board is fully determined. For autonomous driving, the next state in response to the car's action is highly stochastic and noisy. 
In addition, the action of driving (e.g., steering angle and throttle) is continuous. Most model learning algorithms in reinforcement learning only handle discrete actions (e.g. \citet{08uai-linearps,lamapi,rkhsmdp,action_video_prediction_honglak}). 
Recently, CommaAI proposed a simulator model that is learned from real time driving video \cite{comma_ai}. 
\citep{video-predict-2016} used predictive coding inspired by models of the visual cortex to predict the next image in a video sequence without using robot's actions as input.   

In this paper we focus on action models for image prediction, which is of a general form of {\em action-conditioned state prediction} for a specific world representation. 
We provide the first comparison of two recent popular models: the Dynamic Neural Advection (CDNA)  model \citep{video-predict-goodfellow-16} and the feedforward model \citep{ action_video_prediction_honglak}, especially for image prediction on cars. 
On Comma AI driving dataset, 
the CDNA model achieves ${0.9836}$ Structure SIMilarity (SSIM) compared to the ground truth images  while the feedforward model achieves $0.8312$ SSIM. 
To explore the reason of the performance gap between the two models, we conducted a diagnosis experiment. 
Our experiment shows that action tiling encoding is the most important factor leading to the remarkable performance of CDNA regardless of its complicated network architecture. 
We present a light-weight model by action tiling encoding which has a single-decoder feedforward architecture same as \cite{action_video_prediction_honglak}. 
Our model achieves comparable results to CDNA and requires much less memory and computation resources.  
It is especially advantageous to be used on self-driving cars where deploying high-end computation chips such as GPU has to consider cooling, reliability, and cost issues.

\section{Action Models}
{\bfseries Markov Motion Prediction}. 
If a problem is a Markovian Decision Process (MDP), we can characterize it by a tuple, $(\SS, \AA, \PP)$, where $\SS$ is the state space, $\AA$ is the action space, and $\PP$ is a transition kernel. 
The reward in MDP is not within the scope of in this paper and thus left out. 
The problem being Markovian means that the next state after an action only depends on the current state. In particular, the next state $s'$ follows the conditional distribution $\PP(\cdot|s,a)$ after taking $a$ at state $s$. Assume a state representation mapping $\phi: \SS \to T$, where $T$ is a tensor with a generic shape. We interact with the MDP problem and collect a dataset of samples $\{\tp{\phi(s_i), a_i, \phi(s_{i+1})}\}$. The tuple $\tp{\phi(s_i), a_i, \phi(s_{i+1})}$ means that $\phi(s_{i+1})$ is observed immediately after taking action $a_i$ when observing $\phi(s_i)$. For a general interest, we define a model that predicts the expected next state representation, in particular, a model $F$ that minimizes the following Mean-Square-Error (MSE): 
\[
F = \arg\min_{\bold{F}} \sum_{i} \left\| \phi(s_{i+1}) - \bold{F}\left(\phi(s_i), a_i\right) \right\|_2^2.
\]
This model considers only one-step information to predict for the next time step. 
For Markovian problems, this model suffices. 

\begin{figure*}
  \centering
 \includegraphics[width=0.8\linewidth]{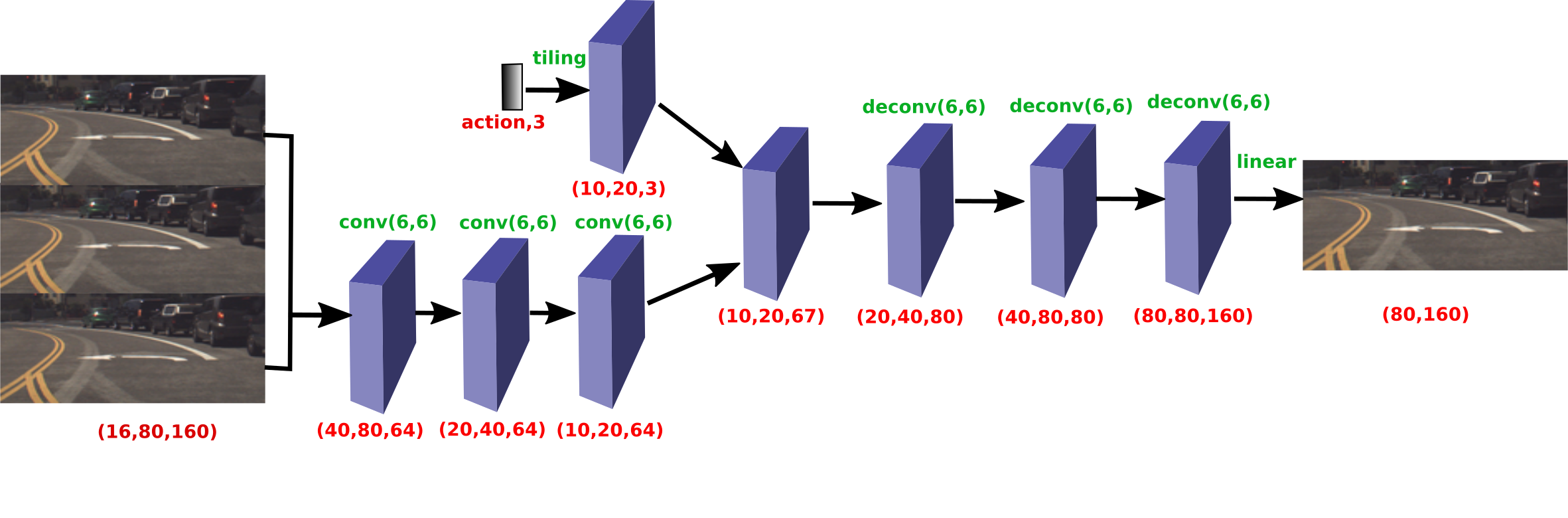}
\caption{Our single-decoder feedforward model with action tiling (SDF-tiling-16) for action-conditioned next image prediction.}
  \label{fig:sdf-tiling}
\end{figure*}

{\bfseries Non-Markov Motion}. 
However, it turns out that for autonomous driving, predicting the next observation of the car is not a Markovian problem. 
Pedestrians and other vehicles can move in an unpredictable way.
However, their behaviors are relatively predictable in a very short time.    
We thus need to extend the model to track the Non-Markov factors:
\begin{equation}\label{eq:Fnon}
F = \arg\min_{\bold{F}} \sum_{i} \left\| \phi(s_{i+1}) - \bold{F}(\phi^h_i,a_i) \right\|_2^2,
\end{equation}
where $\phi^h_i$ is a representation of information up to time step $i$. In practice, $\phi^h_i$ can be a history window of $\phi$ up to the current time step, or a representation recursively constructed from $\phi$ such as in LSTM.  

The model in equation \ref{eq:Fnon} has its various forms in different literature, e.g.,  physics models for motion and contact \cite{physical-interaction-09},
video prediction model for robot arm operation \citep{video-predict-goodfellow-16}, and patch movement in images for car trajectory prediction \citep{visual-predict-rl}.  
In control, it is usually in the form of a set of equations to describe system dynamics \citep{Coates:2008:LCM:1390156.1390175}. In self-driving cars especially motion planning, a specific form of this model (referred to as the {\em Kinematics model} \citep{DBLP:journals/corr/PadenCYYF16}) is used to predict car's motion in a short term. In reinforcement learning, Markovian models and discrete actions are often considered \cite{lamapi,rkhsmdp,action_video_prediction_honglak}. Because all these models capture the effects of actions regardless of their application context, we call them simply the {\em action models}. Action models are specific forms of generative models \citep{goodfellow_generative_2014}. Starting from a given state, one can generate a trajectory of states by applying a sequence of actions. Model-based reinforcement learning algorithms such as Monte Carlo Tree Search (MCTS) rely on action models to generate sample trajectories in order to learn good policies \citep{alphaGo0-2017}.  

This paper is focused on building action models for predicting the next frame. In particular, we focus on $\phi$ being a car camera frame. 

\section{Action-Conditioned Image Prediction}
{\bfseries The SDF Model}. 
The model we consider is the feedforward model for gray-scaled images \cite{action_video_prediction_honglak}. 
The model has a {\em single-decoder feedforward (SDF)} architecture.
Images are encoded using convolution layers and reshaped into a feature vector. 
The action is encoded into a feature vector and combined with the image feature vector into a single vector. 
The resulting vector is reshaped and decoded into the next frame through de-convolution layers. 
The merit of this model is that it is simple in the architecture and easy to understand.  

{\bfseries The CDNA Model}. 
In a recent work \citet{video-predict-goodfellow-16} proposed an image prediction model called Convolutional Dynamic Neural Advection (CDNA) 
that predicts the next frame using a sum of convoluted masks and kernel transformed images. 
The masks and the kernels are learned in two {\em separate decoders} split from a shared representation after a number of convolution layers. 

Another difference between CDNA and SDF is that the history is represented in CDNA using LSTM while in SDF it uses a history window of observations. 
Using LSTM significantly increases the network size. 

The SDF model was shown to predict accurately for video frames in Atari games. 
The CDNA model was shown to predict accurately for robot arm motion, human motion and agent motion in simulators. 
We performed the first comparison for the two models especially for image prediction in driving, giving new insights into the models.

We used Comma AI dataset \footnote{\url{https://github.com/commaai/research}} for evaluation. The training set contains driving logs of images and actions from January 2016 to May 2016. In total there are $459,473$ image-action pairs. Images have been resized into $80\times 160$. 
An action is composed of acceleration, steering angle and brake.
Action controls are normalized with mean and standard deviation calculated from the dataset.
The test set is driving logs in June 2016, totaling $62,953$ image-action pairs. 
We used the Mean-Squared Error (MSE) and SSIM \citep{ssim} for measuring the predicted next images by different models, averaged on the test set.  

\begin{table}
\begin{center}
\begin{tabular}{ l| l| l }
Model& MSE($ 10^{-4}$) & SSIM  \\
\hline
 CDNA \citep{video-predict-goodfellow-16}&  {\bfseries 3.986} & {\bfseries 0.9836}      \\
 SDF \citep{action_video_prediction_honglak}& {\bfseries 23.670} &  {\bfseries 0.8312}\\ 
SDF-recurrent \citep{action_video_prediction_honglak}& 72.600 & 0.6498 \\
Copy-last-frame & 79.20 & 0.6671 \\
\hline
 CDNA-no-current-image & 6.362&  0.9778 \\
 CDNA-no-skip-connection& 4.933 & 0.9798    \\
\hline
 SDF-tiling& 7.050 & 0.9184 \\  
 SDF-tiling-16&  {\bfseries 3.613} & {\bfseries 0.9633} \\  
\end{tabular}
\end{center}
\caption{Image prediction results on Comma AI dataset. }
\label{table:image}
\end{table}

\begin{figure}[htp]
\centering
\includegraphics[width=0.4\textwidth]{{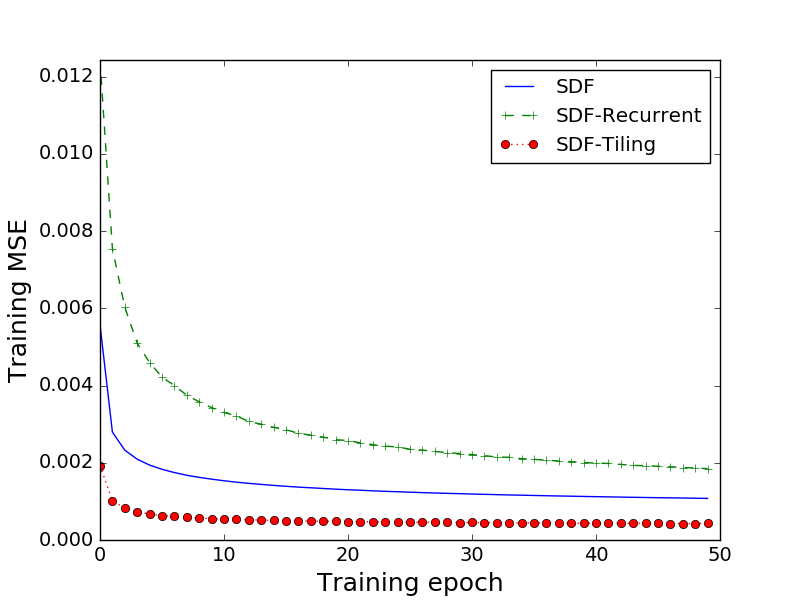}}
\caption{Training errors of SDF and SDF-tiling.}
\label{fig:training-error}
\end{figure}

\begin{figure*}[t!]
\centering
\begin{subfigure}[t]{0.78\textwidth}
\centering
\includegraphics[width=\textwidth]{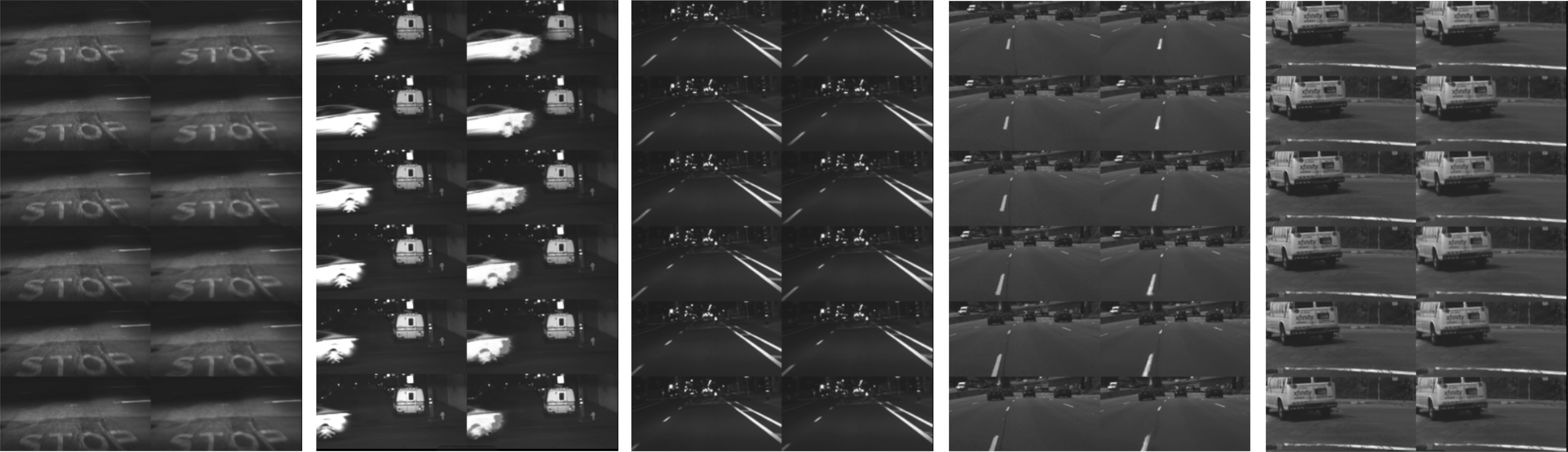}
\caption{CDNA sample predictions.}
\label{fig:cdna}
\end{subfigure}
\hfill
\begin{subfigure}[t]{0.8\textwidth}
\centering
\includegraphics[width=\textwidth]{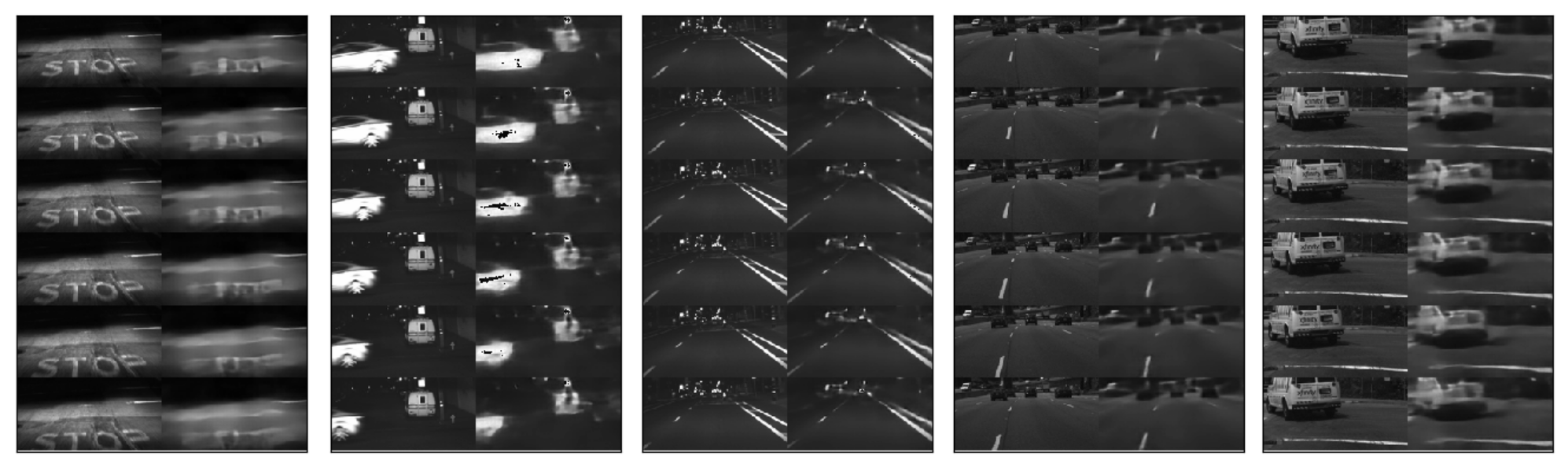}
\caption{SDF sample predictions.}
\label{fig:SDF}
\end{subfigure}
\hfill
\begin{subfigure}[t]{0.8\textwidth}
\centering
\includegraphics[width=\textwidth]{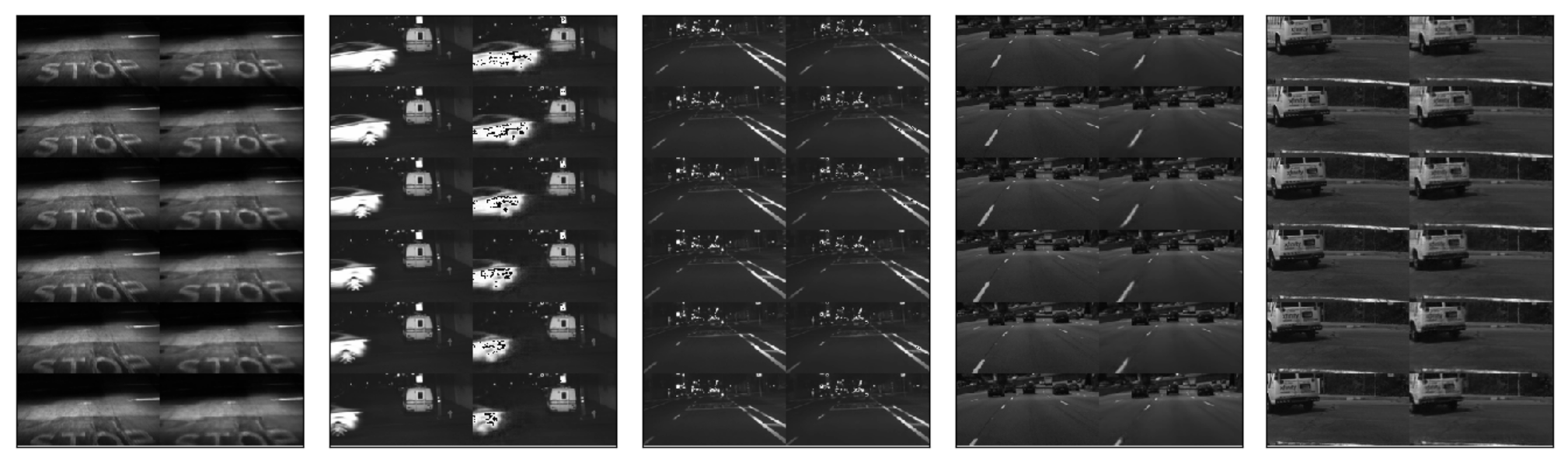}
\caption{SDF-tiling-16 sample predictions.}
\label{fig:sdf-tiling-pred}
\end{subfigure}
\caption{Sample image predictions in the test set. 
In each of the five plots, the left column is the sequence of the ground truth images (running from top to bottom).
The right column is the sequence of predicted next images.  
In a row of each plot, the two images are at the same time step. 
The first plot is a moment when the car passing the stop line.
The second plot is a moment when a car crossing in the front in night. 
The third plot is driving in lane in night with a car in the front. 
The fourth shows a moment switching lanes on highway in daylight.
The fifth shows following a van in the front closing the distance.  
}
\label{fig:image-samples}
\end{figure*}

In our experiment, the CDNA model performed significantly better than the SDF model as shown in Table \ref{table:image}. 
We trained the models on the same training dataset, and optimized their parameters rigorously.  
In the table, CDNA is the original model of \citep{video-predict-goodfellow-16}. 
CDNA achieves a remarkable SSIM of $0.9836$. The predicted images are very sharp as shown in Figure \ref{fig:cdna}. 
The SDF model is the same model as \citep{action_video_prediction_honglak}. 
The model achieved an SSIM score at $0.8312$ and an averaged MSE of $2.367\times 10^{-3}$, which is much inferior to the CDNA model. 
We also run SDF-recurrent, the recurrent model of \citep{action_video_prediction_honglak}, which lead to a lower accuracy than the SDF model in our experiment.

There may have been four factors that have contributed to the outstanding performance of the CDNA model in our experiment:
\begin{itemize}
\item 
The CDNA model employs a special two decoding branches. 
\item 
The model has two skip connections from earlier layers to the mask generation layer. 
\item 
The current image is used as one of the input to the kernel layer. 
\item 
Action is encoded using {\em tiling} instead of vector encoding in the SDF model.
\end{itemize}


\begin{figure*}[htp]
\centering
\begin{subfigure}[t]{\textwidth}
\centering
\includegraphics[width=\textwidth]{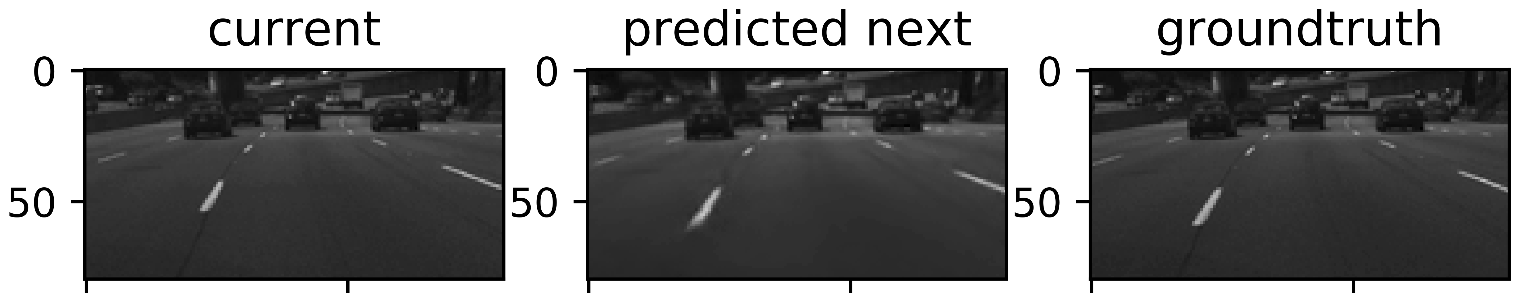}
\end{subfigure}
\vspace{0.3cm}
\begin{subfigure}[t]{\textwidth}
\centering
\includegraphics[width=\textwidth]{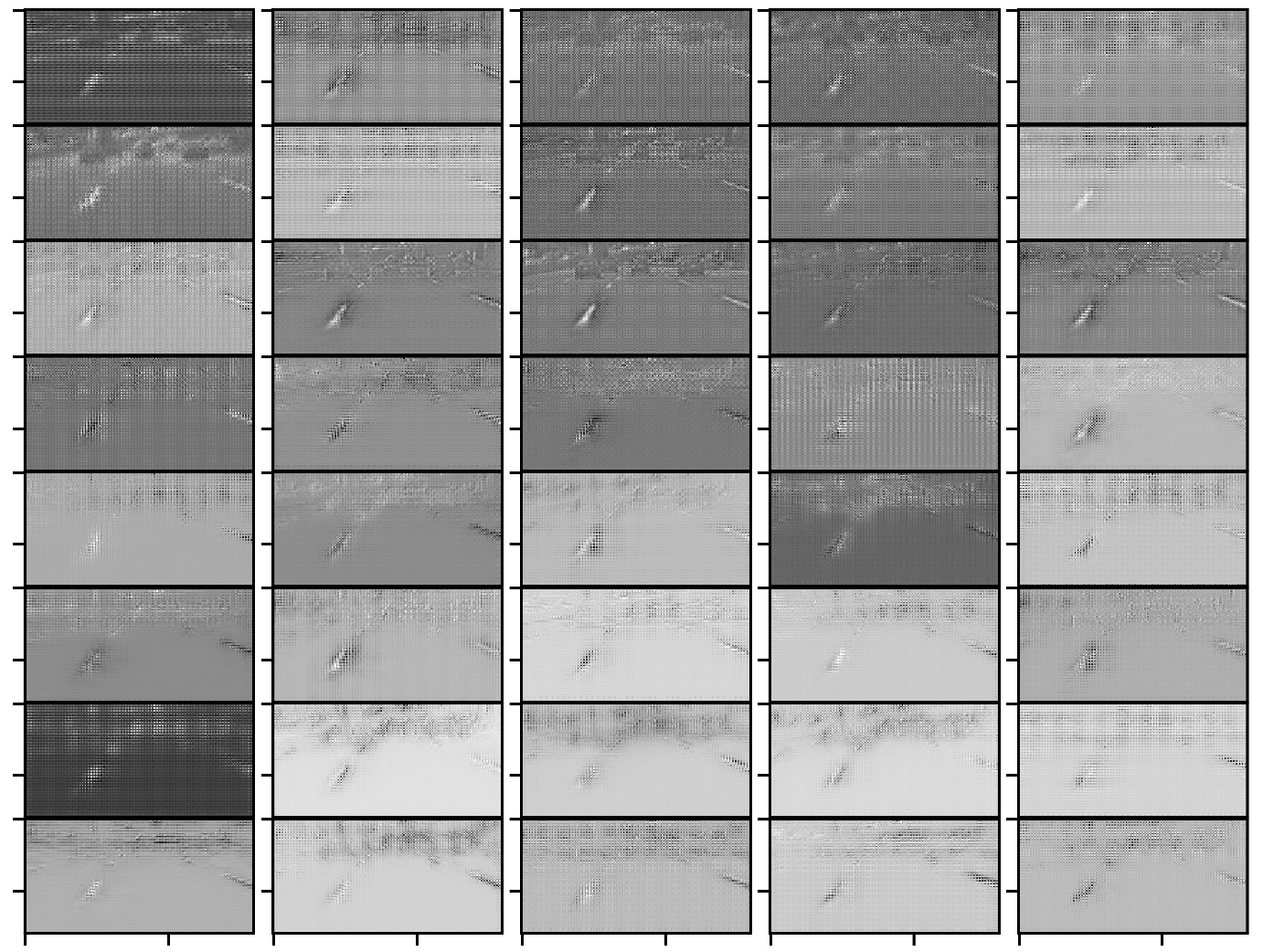}
\end{subfigure}
\caption{Prediction at a sample time step of our model. The top plot shows the current image, the predicted next image and the ground truth. 
The bottom plot shows the $40$ basis images in the last layer for generating the prediction. 
The basis images are presented in the descending order of their weight (from the last layer of our model). 
Thus the weights of these basis images decrease in each row from left to right. 
The top-left basis image has the largest weight. 
}
\label{fig:sample-basis-images}
\end{figure*}

To study the individual effects of these factors, we conducted a diagnosis experiment. 

{\bfseries Effects of Current Image and Skip Connections}.

In particular, {\em CDNA-no-current-image} is the CDNA model leaving out the current image connection in generating one of the transformed images (using its corresponding kernel), achieving an SSIM of $0.9778$.

 {\em CDNA-no-skip-connection} is the CDNA model without the skip connections, getting an SSIM of $0.9798$. 

This shows that leaving out the current image connection to the kernel generation layer or removing the skip connections has  little effects in the performance of CDNA model, degrading only $0.59\%$ and $0.39\%$ respectively.  
  
{\bfseries Effect of Action Tiling}. 
To find whether action tiling was effective, we designed a single-decoder feedforward model with action tiling encoding (called {\em SDF-tiling}), shown in Figure \ref{fig:sdf-tiling}. 
 Our SDF-tiling model achieved a much better performance than SDF, at an SSIM of $0.9184$ and an averaged MSE of $7.050\times 10^{-4}$. 

This shows that tiling encoding of action is much more effective than (dense) vector encoding. 
The reason is that convoluted shapes (e.g., in Figure \ref{fig:sdf-tiling} the convolution layer with shape $(10,20,64)$ before merging with actions) are less tweaked due to 
that action tiling provides a uniform weighting for the shapes.  
The vector encoding of actions introduces a non-uniform weighting to the convolution features, which introduces noises in shapes. 
In addition, the SDF model has four fully connected layers between convolution encoding layers and de-convolution decoding layers (see Figure 9(a) of \citep{action_video_prediction_honglak}). 
This introduces extra noises into the 2D shapes in the output of the convolution encoding layers.     
Altogether the noises from action features and fully connected layers are propagated to de-convolution layers and cause blurry predicted images, as shown in Figure \ref{fig:SDF}.   

{\bfseries Window size}. 
Both the SDF and the SDF-tiling use a window size parameter $4$. 
Objects on the road move much faster than in games and we need a longer history to track them.  
We increased the window size to $16$, and the performance of SDF-tiling on the test set improves to $0.3613\times 10^{-3}$ MSE (better than CDNA) and $0.9633$ SSIM.

{\bfseries Network Size.}
The following table shows the model sizes. 

\begin{tabular}{ l| l }
\hline
Model& \# parameters  \\
\hline
 CDNA \citep{video-predict-goodfellow-16}&  12,661,803    \\
 SDF \citep{action_video_prediction_honglak}& 37,237,825 \\ 
SDF-recurrent \citep{action_video_prediction_honglak}& 70,800,449 \\
 SDF-tiling&   958,400\\  
 SDF-tiling-16&  986,048  \\  
\hline
\end{tabular}

{\bfseries Basis Image Learned}. 
The last layer of SDF-tiling is a linear operation that combines the basis images into the final prediction with trainable weights $w$:
\[
\hat{\phi}(s_{i+1}) = \sum_{i=1}^{n_b} w(i) b(i), 
\] 
where $\bold{b}$ is the collection of the basis images (output of the last second layer).  
In Table \ref{table:image}, the SDF-tiling-16 used the maximum value for $n_b$ which is $80$ according to matrix low rank approximation. 

In practice, using a smaller number of basis images may generalize well too. 
To test this, we run a SDF-tiling with $n_b$ equal to $40$. We also reduced the depth of the other two de-convolution layers to $40$. 
The induced model still gave a descent performance, with an MSE of ${3.940}\times 10^{-4}$ (better than CDNA) and an SSIM of ${0.9576}$ on the test set. 
Figure \ref{fig:sample-basis-images} shows the learned basis images for a sampled time step for this model. 
This model has only ${0.5}$ million parameters. 

{\bfseries Parameters}. 
The best learning rates for training the above models all happen to be $10^{-4}$.  
All the models were trained for $50$ epochs. 

The training error of SDF and our SDF-tiling models are shown in Figure \ref{fig:training-error}. 
The MSE curve of SDF is nearly flat after 40 epochs. 
Using a learning rate of $10^{-3}$ leads to divergence for SDF. 

Our SDF-tiling model used a kernel size of $6$, ADAM optimizer, and RELU activation.

\section{Conclusion}
In this paper, we studied the problem of action-conditioned next image prediction. 
The problem is particularly important for developing model-based reinforcement learning algorithms that can drive a car from raw image observations.
We compared two recent popular models for the next image prediction, especially for predicting the next camera view in driving.  
We found that the CDNA model originally illustrated for robot arm operation \citep{video-predict-goodfellow-16} performed extremely well 
while the feedforward and the recurrent models developed in Atari games setting \citep{action_video_prediction_honglak} failed to give a comparable performance. 
We run diagnosis experiments and found that action-tiling encoding is the most important factor that gives accurate next image predictions. 
Our proposed model combines the best worlds of CDNA and the feedforward model, achieving $0.3613\times 10^{-3}$ MSE (better than CDNA) and $0.9633$ SSIM with a small network size of fewer than one million parameters.





\end{document}